\documentclass{article}
\pdfoutput=1

\usepackage{graphicx} % more modern

\usepackage{natbib}

\usepackage{algorithm}
\usepackage{algorithmic}

\usepackage{tikz}
\usetikzlibrary{bayesnet}

\usepackage{amssymb,amsmath,mathtools}
\usepackage{booktabs}
\usepackage{xspace}

\usepackage[bookmarks=false]{hyperref}
\usepackage[accepted]{icml2013}
\usepackage[noabbrev]{cleveref}

\DeclareMathOperator*{\argmin}{arg\,min}

\renewcommand{\vec}[1]{\mathbf{#1}}
\newcommand{\ten}[1]{\mathbf{#1}} 				% a tensor
\newcommand{\rescal}{\textsc{Rescal}\xspace}
\newcommand{\dist}[1]{\mathcal{#1}}
\newcommand{\from}{\sim}
\newcommand{\normal}[2]{\dist{N}(#1, #2)}	% Normal distribution

\begin{document}

\twocolumn[
\icmltitle{Logistic Tensor Factorization for Multi-Relational Data}
%\icmltitle{On the Order of Data, Tensor Factorizations and\\ Generalization Error Bounds}

% unless you've provided the [accepted] option to the icml2013
\icmlauthor{Maximilian Nickel}{nickel@dbs.ifi.lmu.de}
\icmladdress{Ludwig Maximilian University,
            Oettingenstr. 67, Munich, Germany}
\icmlauthor{Volker Tresp}{volker.tresp@siemens.com}
\icmladdress{Siemens AG, Corporate Technology
            Otto-Hahn-Ring 6, Munich, Germany}

\icmlkeywords{RESCAL, Tensor Factorization, Loss Functions}

\vskip 0.3in
]

%%% Abstract
\begin{abstract}
	Tensor factorizations have become increasingly popular approaches for
	various learning tasks on structured data. In this work, we extend the \rescal
	tensor factorization, which has shown state-of-the-art results for multi-relational
	learning, to account for the binary nature of adjacency tensors. 
	We study the improvements that can be gained via this approach 
	on various benchmark datasets and show that the logistic
	extension can improve the prediction results significantly.
\end{abstract}
%%%

\section{Introduction}
Tensor factorizations have become increasingly popular for learning on various
forms of structured data such as large-scale knowledge bases, time-varying networks 
or recommendation
data~\cite{nickel_factorizing_2012,bordes_learning_2011,bader_temporal_2007,rendle_factorizing_2010}.
The success of tensor methods in these fields is strongly related to their
ability to efficiently model, analyze and predict data with multiple
modalities. Due to their multilinear nature, tensor models overcome limitations
of linear models, such as their limited expressiveness, but at the same time
remain more scalable and easier to handle then general non-linear approaches.

\rescal~\cite{nickel_three-way_2011,nickel_factorizing_2012} is a tensor factorization 
for dyadic multi-relational data which has been shown to achieve
state-of-the-art results for various relational learning tasks such as link
prediction, entity resolution or link-based clustering. Briefly, the \rescal
model can be summarized as following: For relational data
with $K$ different dyadic relations and $N$ entities, a third-order
\emph{adjacency tensor} $\ten{X}$ of size  $N \times N \times K$ is created, where
\[
	x_{ijk} = \begin{cases}
		1, & \text{if } Rel_k(Entity_i, Entity_j) \text{ is true}\\
		0, & \text{otherwise.}
	\end{cases}
\]
This adjacency tensor $\ten{X}$ is then factorized into latent representations of entities
and relations, such that
\[
	X_k \approx A R_k A^T
\]
where $X_k$ is the $k$-th frontal slice of $\ten{X}$. After computing the
factorization, the matrix $A \in
\mathbb{R}^{N \times r}$ then holds the latent representations for the entities 
in the data, i.e.~the row $\vec{a}_i$ holds the latent representation of the 
$i$-th entity. Furthermore, $R_k \in \mathbb{R}^{r \times r}$ can be
regarded as the latent representation of the $k$-th predicate, whose entries 
encode how the latent components interact for a specific relation.
Since $R_k$ is a \emph{full, asymmetric} matrix, the factorization can also 
handle directed relations. 
When learning the latent representation of an entity, unique global 
representation allows the model to efficiently access information that is 
more distant in the relational graph via information propagation through 
the latent variables.
%An important property  \rescal for \emph{relational} learning 
%is that entities have a unique global representation via the latent factor $A$. 
%During learning the latent representation of an entity, this allows the model 
%to efficiently access information that is more distant in the relational graph
%via information propagation through the latent variables. 
For instance, it has been shown that \rescal can propagate information about party
membership of presidents and vice presidents over multiple relations, 
such that the correct latent representations are learned even when the party
membership is unknown~\cite{nickel_three-way_2011}. Moreover, since the entries of $\ten{X}$ are 
mutually independent given the latent factors $A$ and $\ten{R}_k$, 
prediction is very fast, as it reduces to simple vector-matrix-vector
products.

In its original form, the \rescal factorization is computed by 
minimizing the least-squares error 
between the observed and the predicted entries; in a probabilistic
interpretation this implies 
that the random variation of the data follows a Gaussian distribution, i.e.~that
\[
	x_{ijk} \from \normal{\theta}{\sigma^2}
\]
where $\theta$ are the parameters of the factorization.
However, a Bernoulli is more appropriate for binary variables with
\[
	x_{ijk} \from Bernoulli(\theta)
\]
where the parameter $\theta$ is again computed via the factorization of the
corresponding adjacency tensor. In the following, we will present a learning 
algorithm based on
logistic regression\footnote{In the theory of the exponential family, the
	logistic function describes the inverse parameter mapping for the Bernoulli
distribution} using the Bernoulli likelihood model and evaluate on benchmark data what gains can be expected
from this updated model on relational data.

\section{Methods}
\begin{figure}[bt]
	\centering
	\begin{tikzpicture}
  		% Define nodes
		\tikzstyle{mynode} = [inner sep=4pt, font=\normalsize,thick]
  		\node[obs,mynode] (x) {$x_{ijk}$};
		\node[const,above=2.75cm of x] (sigA) {$\sigma_A$};
		\node[const,right=of x, yshift=-2cm] (sig) {$\sigma$};
  		\node[latent,mynode,above=of x, xshift=-1.75cm](ai) {\normalsize$\vec{a}_i$};
  		\node[latent,mynode,above=of x, xshift=1.75cm] (aj) {$\vec{a}_j$};
  		\node[latent,mynode,below=1.5cm of x] (R) {$R_k$};
		\node[const,left=of R] (sigR) {$\sigma_R$};

  		% Connect the nodes
  		\edge[thick]{ai,aj,R} {x} ; %
		\edge[thick] {sigA} {ai,aj} ; %
		\edge[thick] {sigR} {R} ; %
		\edge[thick,dotted] {sig} {x} ; %

  		% Plates
  		\plate[thick] {ajx} {(aj)(x)} {$N$} ;
  		\rplate[thick] {aix} {(ai)(x)(ajx.north west)(ajx.south west)} {$N$} ;
		\plate[thick] {} {(x)(R)(ajx.south west)(aix.south east)(ajx.west)} {$K$};
	\end{tikzpicture}
	\caption{Graphical model in plate notation of the \rescal factorization. 
		The parameter $\sigma$ is only
		present when the random variable $x_{ijk}$ follows a Normal distribution.}
	\label{fig:gm}

\end{figure}
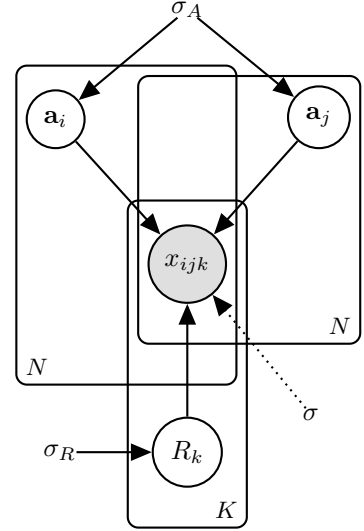

In the following, we interpret \rescal from a
probabilistic point of view. Each entry $x_{ijk}$ in $\ten{X}$ is regarded as a 
random variable and we seek to compute the MAP estimates of $A$ and $\ten{R}$
for the joint distribution
\begin{equation}
	p(\ten{X}|A,\ten{R}) = \prod_{ijk}p(x_{ijk}|\vec{a}^T_iR_k\vec{a}_j).
	\label{eq:rescal-probal}
\end{equation}
\Cref{fig:gm} also shows the graphical model in plate notation for the
factorization.
We will also fix the prior distributions of the latent factors to the Normal
distribution, i.e. we set
\begin{align*}
	\vec{a}_i & \from \normal{0}{\lambda_A I}\\
	R_k & \from \normal{0}{\lambda_R I}
\end{align*}
Furthermore, we will maximize the log-likelihood of~\cref{eq:rescal-probal},
such that the general form of the objective function that we seek to 
optimize is
\begin{equation}
	\argmin_{A,\ten{R}} \mathrm{loss}(X; A, \ten{R}) + 
		\lambda_A\|A\|_F^2 +
		\sum_k \lambda_R\|R_k\|_F^2
	\label{eq:gen}
\end{equation}
The nature of the loss function depends on the distribution that we assume for
$x_{ijk}$. In the following we consider the least-squares and the logistic loss
function.

\subsection{Least-Squares Regression}
In its original form, \rescal sets the loss function to
\begin{equation}
	\mathrm{loss}(X; A, \ten{R}) \coloneqq \sum_k \|X_k - AR_kA^T\|_F^2.
	\label{eq:rescal-als}
\end{equation}
In this case, \cref{eq:gen} and \cref{eq:rescal-als}
maximize the log-likelihood of~\cref{eq:rescal-probal} when
\begin{align*}
	x_{ijk} & \from \normal{\vec{a}_i^TR_k\vec{a}_j}{\sigma^2}
\end{align*}
It should be noted that although the least-squares error does not imply the 
correct error model, it has the appealing property that it enables a 
very efficient and scalable implementation. 
An algorithm based on alternating least-squares updates of the
factor matrices, has been shown to scale up to large knowledge bases via
exploiting the sparsity of relational data. For instance, it has been used to factorize
YAGO, an ontology which consists of around 3 million entities, 40 relations, and 
70 million known facts on a single desktop
computer~\cite{nickel_factorizing_2012}. In the following we will
refer to this implementation as \rescal-ALS.

\subsection{Logistic Regression}
To describe the random variation in the data via a Bernoulli distribution, we
set
\begin{multline}
	\mathrm{loss}(\ten{X}; A, \ten{R}) \coloneqq \\
	-\sum_{ijk} x_{ijk} \log \sigma(\theta_{ijk}) 
	+ (1-x_{ijk}) \log \left(1 - \sigma(\theta_{ijk})\right)
	\label{eq:rescal-logit}
\end{multline}
where 
\[
	\sigma(\theta_{ijk}) = \frac{1}{1 + \mathrm{exp}(-\vec{a}_i^TR_k\vec{a}_j)}
\]
Now, \cref{eq:gen} and \cref{eq:rescal-als}
maximize the log-likelihood of~\cref{eq:rescal-probal} when
\begin{align*}
	x_{ijk} & \from Bernoulli(\sigma(\theta_{ijk}))
\end{align*}
Since, there exists no closed form solution to compute
\cref{eq:rescal-logit}, we use a gradient based approach
to compute~\cref{eq:rescal-logit} via quasi-Newton optimization, i.e.~via the L-BFGS
algorithm. The partial gradients for $A$ and $R_k$ are
\begin{align*}
	\frac{\partial}{\partial A} = & \sum_k \left[\sigma(AR_kA^T) -
		X_k)\right]AR_k^T +\\
		& \quad\quad\quad \left[\sigma(AR_kA^T) - X_k)\right]^TAR_k + 2\lambda_AA \\
	\frac{\partial}{\partial R_k} = & A^T \left[\sigma(AR_kA^T) - X_k\right] A +
	2\lambda_RR_k
\end{align*}
where $\sigma(AR_kA^T)$ denotes the elementwise application of
$\sigma(\cdot)$ to $AR_kA^T$.
Unfortunately, terms of the form 
\[
	\left[\sigma\left(AR_kA^T\right) - X_k\right]A
\]
can not be reduced to a significantly simpler form, due to the logistic function. Hence, this
approach currently requires to compute the dense matrix $AR_kA^T$, what limits its
scalability compared to the alternating least-squares approach. In the
following, we will refer to this approach as \rescal-Logit

%\subsection{Non-Negative Factorization}
%At last we also consider a non-negative variant of the \rescal factorization.
%Non-Negative Matrix Factorization~(NMF) is a popular approach, e.g.~for NLP
%tasks. 
%\begin{align*}
%	& \argmin_{A,\ten{R}} & & \sum_k \|X_k - AR_kA^T\|_F^2 + \lambda\|A\|_1 +
%	\mu\|R_i\|^2\\
%	& \text{subject to} & & a_{ij}, r_{ijk} \geq 0
%\end{align*}

%\subsection{Sigmoidal Transfer Function}
%To obtain valid probabilities for the least-squares cost-function, we also
%consider the following method
%\[
%	\theta_{ijk} = \sigma_{\epsilon}(\vec{a}_i^T R_k \vec{a}_j)
%\]
%where 
%\[
%	\sigma_{\epsilon}(x) =
%	\begin{cases}
%		x & \text{if } \epsilon < x < 1 - \epsilon \\
%		\frac{\epsilon}{e}\mathrm{exp}({\frac{x}{\epsilon}}) & \text{if } x \leq \epsilon \\
%		1 - \frac{\epsilon}{e}\mathrm{exp}({\frac{1 - x}{\epsilon}}) & \text{if } x \geq 1 - \epsilon \\
%
%	\end{cases}
%\]
%Note that such a postprocessing to obtain valid probabilities has been motivated
%in [7]. Since $\sigma_{\epsilon}$ is monotonic, the ordering of the
%probabilities is maintained. With $\epsilon \rightarrow 0$ we obtain the correct posterior probabilities in the asymptotic case.

\section{Experiments}
\begin{table}[tb]
	\centering
	\caption{Evaluation results of the area under the precision-recall curve on
	the Kinships, Nations, Presidents, and Bacteriome datasets.}
	\begin{tabular}{lcccc}
		\toprule
		& Kinships & Nations & Pres. & Bact.\\
		\midrule
		\rescal-ALS & 0.966 & 0.848 & 0.805 & 0.927\\
		\rescal-Logit & \textbf{0.981} & 0.851 & 0.800 & \textbf{0.938}\\
		MLN & 0.85 & 0.75 & - & -\\
		IRM & 0.66 & 0.75 & - & -\\
		\bottomrule
	\end{tabular}
	\label{tab:results}
\end{table}
To evaluate the logistic extension of \rescal, we conducted link-prediction
experiments on the following datasets:
\begin{description}
	\item[Presidents] Multi-relational data, consisting of
		presidents of the United States, their vice-presidents as well as the parties
		of presidents and vice presidents. 
	\item[Kinships] Multi-relational data, consisting of several kinship
		relations within the Alwayarra tribe. 
	\item[Nations] Multi-relational, data consisting of relations between nations
		such as treaties, military actions, immigration etc.
	\item[Bacteriome] Uni-Relational data, consisting of protein-protein and functional interactions within the context of an E. coli knowledgebase
\end{description}

For all datasets we performed 10-fold cross-validation and evaluated the results
using the area under the precision-recall curve.
In case of the presidents data, the task was to predict the party membership for
presidents only based on the party memberships of their vice-presidents (and vice-versa).
For all other datasets, cross-validation has been applied over all existing relations.
It can be seen from the results in~\cref{tab:results} that the logistic
extension of \rescal can considerably improve the prediction results.
Especially the improvements for Kinships and Bacteriome are noteworthy,
considering the already very good results of \rescal-ALS.

\section{Conclusion}
To improve the modeling of multi-relational data, we have presented
an extension for \rescal based on logistic regression. We have shown on several
benchmark datasets that the logistic
extension can improve the prediction results significantly.
While the evaluation results are very encouraging, future work will have to
address the scalability of the presented approach, as the scalability of its 
current implementation is too limited for practical use on larger datasets.

{
	\footnotesize
	\subsubsection*{Note added in proof}
	Independently, a similar logistic extension of the \rescal factorization 
	has been proposed in \cite{london_multi-relational_2013}.

	\bibliography{SLG2013}
	\bibliographystyle{icml2013}
}
\end{document}